# Corn Yield Prediction based on Remotely Sensed Variables Using Variational Autoencoder and Multiple Instance Regression

Zeyu Cao, Yuchi Ma, Zhou Zhang, *Member*, *IEEE*

*Abstract*— In the U.S., corn is the most produced crop and has been an essential part of the American diet. To meet the demand for supply chain management and regional food security, accurate and timely large-scale corn yield prediction is attracting more attention in precision agriculture. Recently, remote sensing technology and machine learning methods have been widely explored for crop yield prediction. Currently, most county-level yield prediction models use county-level mean variables for prediction, ignoring much detailed information. Moreover, inconsistent spatial resolution between crop area and satellite sensors results in mixed pixels, which may decrease the prediction accuracy. Only a few works have addressed the mixed pixels problem in large-scale crop yield prediction. To address the information loss and mixed pixels problem, we developed a variational autoencoder (VAE) based multiple instance regression (MIR) model for large-scaled corn yield prediction. We use all unlabeled data to train a VAE and the well-trained VAE for anomaly detection. As a preprocess method, anomaly detection can help MIR find a better representation of every bag than traditional MIR methods, thus better performing in large-scale corn yield prediction. Our experiments showed that variational autoencoder based multiple instance regression (VAEMIR) outperformed all baseline methods in large-scale corn yield prediction. Though a suitable meta parameter is required, VAEMIR shows excellent potential in feature learning and extraction for large-scale corn yield prediction.

## I. INTRODUCTION

CORN is the most produced crop in the U.S. and makes up a significant part of the American diet [1]. In the Midwestern states, known as the U.S. corn belt, over 15 billion bushels of corn were produced in 2021 [2]. Accurate and timely estimation of corn yield in the U.S. is therefore of great importance for farm resource management, food security monitoring, and market planning [3]. As a result, corn yield prediction is appealing more and more attention in precision agriculture.

A traditional yield prediction method is survey-based, which is time-consuming and labor-intensive [4]. Also, some yield prediction methods are adopting physical simulation models, which require management-related input parameters that are difficult to obtain [5]. Recently, with the support of remote sensing technology, more remote sensing variables based methods have been proposed for large-scale yield prediction. Remote sensing is a tool to monitor the earth's resources using space technologies for higher precision and accuracy [6]. The typical responses of the targets are different in the remote sensing observations, so it is possible to use them to distinguish the vegetation, bare soil, water and other similar features. Besides, remote sensing technology can collect data without physical contact with the crop, enabling large-scale observations and analysis. Usually, informative features are extracted from satellite images for monitoring crop growth conditions, such as vegetation indices (VIs). With the help of proper machine learning (ML) models, the estimation of crop yield can be obtained by these features. Researchers have proposed various ML methods for corn yield prediction. For example, Khaki et al. [7] designed a deep neural network (DNN) for crop yield prediction based on weather and soil variables. Nevavuori et al. [8] adopted convolutional neural networks (CNN) for crop yield prediction using remote sensing variables. To consider the time dependencies of environmental factors, Khaki et al. [9] combined a convolutional neural network with a recurrent neural network (CNN-RNN) for crop yield prediction. Maimaitijiang et al. [10] utilized DNN to process multisource data for soybean yield prediction. Schwalbert et al. [11] integrated different ML algorithms for soybean yield prediction using weather and satellite data. More recently, Ma et al. [12] developed a bayesian neural network (BNN) to predict the corn yield and quantify the predictive uncertainty.

Though the development in ML for crop yield prediction is prosperous, there are still some obstacles in large-scale corn yield prediction. For county-level yield prediction, most of the previous studies spatially aggregate the data to county-level [12], [13], which causes the loss of detailed information. Moreover, inconsistent spatial resolution between crop fields and satellite sensors results in mixed pixels, which makes it more difficult to extract accurate VIs for all crop fields. Accurate yield prediction models can only be obtained once these two problems are addressed. Multiple instance regression (MIR) [14] is a promising method for utilizing more detailed information. In county-level crop yield prediction, MIR considers different counties as bags, and each bag contains some instances, such as remote sensing pixels. Each pixel is

Manuscript received xxx. This research was supported by USDA National Institute of Food and Agriculture, Agriculture and Food Research Initiative project 1028199; and the University of Wisconsin–Madison, Office of the Vice Chancellor for Research and Graduate Education with funding from the Wisconsin Alumni Research Foundation.

Zeyu Cao, Yuchi Ma, Zhou Zhang are with Department of Biological Systems Engineering, University of Wisconsin-Madison, Madison, WI 53706, USA (corresponding author: Zhou Zhang, email:zzhang347@wisc.edu).

composed of some variables, such as VIs and climate variables. In practice, using every pixel in a county for regression is unacceptable in time and computation, so instances are usually subsampled from all pixels in a county. When the number of instances is enough, these instances can reflect all information we need for a county. Finding a proper representation by multiple instances is essential in MIR, and many related works have been done. For example, Prime-MIR [15] assumes a primary instance exists in each bag, and cluster-MIR [16] was proposed to process structured data. However, few MIR methods take the mixed pixels into consideration, so mixed pixels can still decrease the performance of MIR in large-scaled corn yield prediction.

To avoid the influence of mixed pixels, we adopt a variational autoencoder (VAE) [17] structure to improve the prediction accuracy. VAE is a wide-used neural network structure for unsupervised feature learning. It aims to map the original input to a latent vector with a lower dimension, then reconstructs the input from the latent vector. This process is unsupervised because reconstruction error (e.g., Euclidean distances between output and input) can be used to optimize the network's parameters. For a well-trained VAE, it can also be used for anomaly detection [18]. We propose a corn yield prediction method combining variational autoencoder with multiple instance regression (VAEMIR) in this letter. We use VAE for anomaly pixels detection, and data without anomaly pixels are used for corn yield prediction in the MIR way. The main contributions of this work are listed as follows:

1) We adopt the MIR method for large-scale corn yield prediction, which utilizes more detailed information than using the mean of counties.
2) We introduce a VAE structure to reduce the influence caused by mixed pixels in corn yield prediction.

## II. METHODOLOGY

### A. Variational Autoencoder (VAE) for Anomaly Detection

As stated above, inconsistent spatial resolution between crop fields and satellite sensors can cause mixed pixels in the satellite imagery. Usually, the spatial resolution of the satellite is over one hundred meters, e.g., 500m. However, the spatial resolution of crop fields may vary. Large and continuous crop fields' spatial resolution can be larger than 500m. Such areas become pure pixels in the satellite imagery. The definition of pure pixels is that only one spectrally unique material exists on the earth.

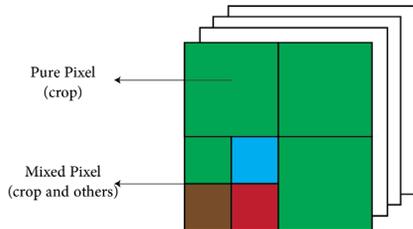

Fig. 1. Illustration of pure pixel and mixed pixel in a satellite imagery.

As shown in Fig 1, green areas are corresponded to crop fields on the earth. Large crop fields can be a pure pixel in the satellite imagery. For those crop fields with smaller spatial resolution, e.g., 30m, they may neighbor various surface features, such as bare soil and pool. Then these features under the 500m scale are aggregated into a single pixel in the satellite imagery, a mixed pixel. VIs cannot reflect real crop growth status if they are computed from mixed pixels because the existence of other surface materials may affect the spectral values in the mixed pixels. As a result, remote sensing variables derived from mixed pixels can be seen as anomaly data. Furthermore, that is why we adopt a variational autoencoder for anomaly detection.

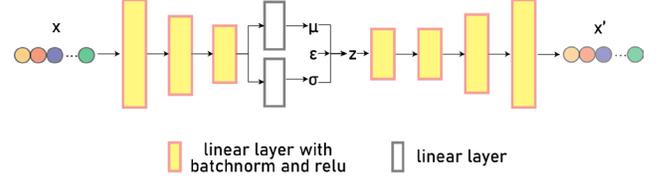

Fig. 2. Structure of VAE.

As Fig 2 shows, Variational Autoencoder (VAE) is an artificial neural network architecture widely used for unsupervised dimensionality reduction, feature extraction and content generation. VAE is composed of an encoder and a decoder. The encoder maps the input x to a latent code z with a lower dimension, and the decoder reconstructs the input x from z. The output of VAE is reconstructed input x', by reducing the distance between x and x', VAE is well-trained for many downstream tasks.

VAE encodes the input to two variables, μ and σ. Assume ε is a random value drawn from the normal distribution. Then z can be obtained by Eq. (1). In the training process, two loss functions are utilized to optimize the VAE, just as shown in Eq. (2), (3), (4), N means the number of samples. Eq. (2) is latent loss, which controls the distribution of z. Moreover, Eq. (3) is a reconstruction loss that controls the distance between input and output.

$$z = \mu + \varepsilon \times \sigma \quad (1)$$

$$\mathcal{L}_{\text{latent}} = \frac{1}{2} \sum_{i=1}^{N} \left( \mu_{(i)}^2 + \sigma_{(i)}^2 - \log \sigma_{(i)}^2 - 1 \right) \quad (2)$$

$$\mathcal{L}_{recon} = \sqrt{\sum_{i=1}^{N}(x_i - x_i')^2} \quad (3)$$

$$\mathcal{L} = \mathcal{L}_{\text{latent}} + \mathcal{L}_{\text{recon}} \quad (4)$$

Assuming there are many normal samples and a few anomaly samples in the dataset, a well-trained VAE tends to get a larger reconstruction loss between anomaly input and output because it is trained to fit the distribution of many normal samples. As a result, we can use a well-trained VAE for anomaly detection based on the reconstruction loss.

### B. VAE-based Multiple Instance Regression (VAEMIR)

To better describe the county-level growth status of corn, Multiple Instance Regression (MIR) subsamples N pixels instead of using the mean of all pixels in each county. A pixel comprises D variables, such as VIs, soil properties and temperature information. Also, a pixel can be named as an instance in MIR. N instances compose a bag, which corresponds to a county. A bag with N instances can provide detailed information about a county because it has the same data distribution as the county if N is big enough. With N instances provided, MIR aims to find the best representation (prototype in VAEMIR) for each bag.

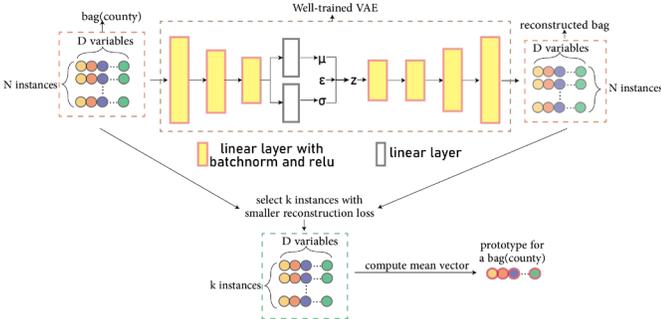

Fig. 3. Illustration of VAEMIR.

The flowchart of VAEMIR is shown in Fig. 3. Assume we have a well-trained VAE, and we can use it to calculate the reconstruction loss of every instance in each bag. The reconstruction loss is shown in Eq. (3). Then, we select k instances with smaller reconstruction loss to represent the input bag, and k is a preset parameter. The unselected instances are removed because they are more likely to be anomaly instances. Then we compute the mean vector of the k instances to get the prototype for the bag. This way, we can get prototypes for all bags and use these prototypes for regression.

## III. EXPERIMENTAL SETUP

### A. Study Area and Data Acquisition

We use 12 states in the U.S. corn belt as our study area, including North Dakota, South Dakota, Minnesota, Wisconsin, Iowa, Illinois, Indiana, Ohio, Missouri, Kansas, Nebraska and Michigan. We build a dataset using remote sensing imagery, weather observations and soil properties to get better prediction performance. Specifically, we use three complementary VIs, including Green Chlorophyll Index (GCI), Enhanced Vegetation Index (EVI), and Normalized Difference Water Index (NDWI). They are all extracted from the MODIS MCD43A4 product [19] with 500m spatial resolution. For weather observations, we collected daily mean air temperature (Tmean), maximum air temperature (Tmax), and total precipitation (PPT) from the Parameter elevation Regressions on Independent Slopes Model (PRISM) dataset with 4km spatial resolution. Moreover, daytime and nighttime land surface temperature (LSTday and LSTnight) were extracted from the MODIS MYD11A2 product [20] with a 1km spatial resolution. For soil properties, we use Available Water Holding Capacity (AWC), Soil Organic Matter (SOM) and Cation Exchange Capacity (CEC) from Soil Survey Geographic database (SSURGO) [21] with 30m spatial resolution.

To extract the corn-specific features, we use the USDA National Agricultural Statistics Service (NASS) Cropland Data Layer (CDL) as the crop mask to mask out areas on non-corn lands [22] with 30m spatial resolution. All variables are concatenated pixel by pixel. Then we randomly subsample 100 pixels with 30m spatial resolution in each county.

Time-series remote sensing and weather variables were aggregated temporally to a 16-day interval from the middle of April to early October to cover the corn-growing season to reduce the computation complexity. County-level historical yield records were collected from the USDA NASS [22] for model development and evaluation. Finally, the five-year historical average yield and current year are concatenated to processed pixels.

After all the processes, we get a corn yield prediction dataset ranging from 2008 to 2021 and have around 750 bags per year. Each bag is composed of 100 instances, and each is a 159-dimensional feature.

### B. Model Training and Evaluation

To illustrate the utility of VAEMIR, we use four baseline methods for comparison, including Instance-MIR [23], Mean Regression, Prime-MIR [15], and Cluster-MIR [16]. Instance-MIR treats every instance as a sample when training the regressor, just like no bags. Mean Regression uses the mean vector of all instances in a bag for regression. Prime-MIR selects one most representative instance in each bag for regression. Cluster-MIR divides all instances into clusters to get many regressors, then uses the most representative regressor for all bags. We use a multi-perceptron regressor composed of two hidden layers for regression in all methods, including VAEMIR. Hidden layers are composed of 128 and 64 neurons.

All models are evaluated in six testing years, 2016-2021, while data in all preceding years since 2008 are used for model training for each testing year. For example, when testing in 2016, data samples from 2008-2015 were used for model training. Two metrics, i.e., root mean square error (RMSE) and coefficient of determination ($R^2$), are used to evaluate the model performance. All experiments are repeated five times, and the mean values of all results are recorded.

## IV. RESULTS AND DISCUSSION

### A. Model comparison

The model evaluation results in each testing year are shown in Table 1. For every method, we recorded their RMSE and $R^2$ each year, then calculated the mean value for comparison. For VARMIR, we additionally recorded the preset k value in order to find the range of best k. The best results in each year are highlighted in bold.

Instance-MIR directly uses every instance for regression, ignoring the bag structure. In 2016 and 2021, Instance-MIR got lower $R^2$ and RMSE than all methods, showing that equally treating every instance may decrease the prediction performance. In other words, MIR requires proper instances selection strategy for better performance.

However, it seems strange that Prime-MIR and Cluster-MIR cannot outperform Mean Regression in all testing years. Especially, Prime-MIR got extremely low $R^2$ and RMSE in 2019 and 2020, as Cluster-MIR did in 2018 and 2020.

TABLE I
MODEL EVALUATION RESULTS IN 2016-2021.

| Year | Instance-MIR | | Mean Regression | | Prime-MIR | | Cluster-MIR | | **VAEMIR** | | |
|---|---|---|---|---|---|---|---|---|---|---|---|
| | RMSE | $R^2$ | RMSE | $R^2$ | RMSE | $R^2$ | RMSE | $R^2$ | RMSE | $R^2$ | k |
| 2016 | 1.23 | 0.61 | 1.01 | 0.74 | 1.04 | 0.72 | 1.05 | 0.71 | **0.96** | **0.76** | 65 |
| 2017 | 1.02 | 0.79 | 1.03 | 0.78 | 1.08 | 0.77 | 1.21 | 0.70 | **0.94** | **0.82** | 60 |
| 2018 | 1.35 | 0.65 | 1.34 | 0.66 | 1.36 | 0.65 | 1.84 | 0.36 | **1.17** | **0.74** | 70 |
| 2019 | 1.22 | 0.58 | 0.89 | 0.78 | 1.23 | 0.57 | 1.16 | 0.62 | **0.87** | **0.79** | 80 |
| 2020 | 1.19 | 0.65 | 1.10 | 0.70 | 1.46 | 0.47 | 1.50 | 0.44 | **1.05** | **0.72** | 80 |
| 2021 | 1.62 | 0.64 | 1.19 | 0.80 | 1.59 | 0.65 | 1.55 | 0.67 | **1.15** | **0.82** | 70 |
| Average | 1.27 | 0.65 | 1.09 | 0.74 | 1.29 | 0.64 | 1.38 | 0.59 | **1.02** | **0.78** | / |

Those are evidence that large-scale corn yield prediction is hard for traditional MIR methods because they are not designed for large-scale scenes. The data variance among different states may harm the feature learning process. As a result, Mean Regression turns out to be more suitable for such large-scale scene because calculating average value can lower the variance among instances and perform better than traditional MIR methods.

Unlike baseline MIR methods, VAEMIR outperforms all methods in all testing years in terms of both $R^2$ and RMSE, including Mean Regression. For large-scale corn yield prediction, variance among different states and noise induced by mixed pixels can both decrease the prediction performance. Though Mean Regression can decrease the influence of regional variance, it can still be affected by mixed pixels. Unlike other methods, VAEMIR uses a well-trained VAE for anomaly detection, which can filter instances with high variance. In this way, VAEMIR can decrease the influence of regional variance and mixed pixels in the meantime. Therefore, VAEMIR outperforms Mean Regression and traditional MIR methods in all testing years.

Besides regional variance, there is also temporal variance among different testing years, causing some years to be hard to predict, such as 2018. None of the baseline methods can get $R^2$ higher than 0.7, while VAEMIR got an $R^2$ of 0.74, which shows VAEMIR's ability to reduce the influence of temporal variance. Moreover, VAEMIR got the best average result of all testing years, which also shows that VAEMIR is more robust than Mean Regression and traditional MIR methods among different testing years.

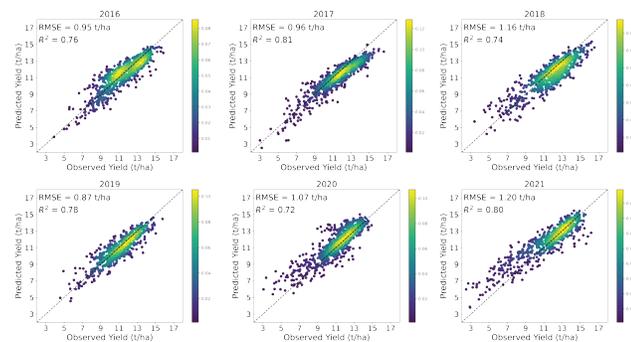

Fig. 4. Agreement of predicted yield and reported yield (VAEMIR).

For better evaluation of VAEMIR, we showed the agreement of predicted yield and reported yield by combining results of VAEMIR in all testing years (Fig. 4). Every point in the figures represents a county (bag), and the location of points reflects the prediction performance of regression models. These data are overestimated if the points are under the dashed line. These data are underestimated if the points are above the dashed line. Reviewing the agreement figures shows that the overestimated and underestimated points are relatively evenly distributed on both sides of the dashed line, which means that VAEMIR does not have a severe bias in prediction.

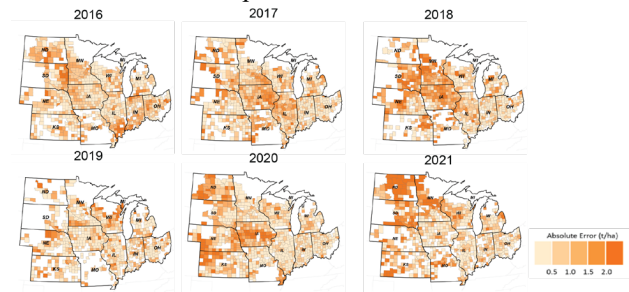

Fig. 5. Absolute error in different states (VAEMIR).

Fig. 5 shows the absolute error of VAEMIR's prediction in different states. Every pixel block represents a county in a state; deeper color means a more significant absolute error in prediction. We can find that most states have no prominent error clusters by analyzing the absolute error figures. In other words, VAEMIR can avoid the influence of the spatial variance among different states and learn more general features for prediction.

### B. Model Performance with Different k

In VAEMIR, k is a preset parameter that can affect the performance of the prediction. To better evaluate VAEMIR, we recorded the prediction results with different k in Fig. 6.

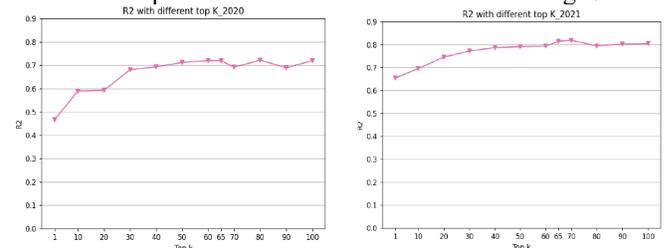

Fig. 6. Model Performance with different k (VAEMIR).

As shown in Fig. 6, the $R^2$ of the regressors will change when the k varies. The curves in 2020 and 2021 clearly show the typical relationship between k and $R^2$ in VAEMIR. When k is set to 1, VAEMIR degrades to Prime-MIR because only one instance is selected as the bag prototype. The assumption in Prime-MIR is not suitable for large-scale corn yield prediction because it is nearly impossible to use only one instance to represent the crop growth status in such a large area. As a result, regressors always get the worst prediction performance when k is set to 1. When k is set to 100, i.e., k equals N, it means all instances in a bag are used to compute the prototype. Then VAEMIR degrades to Mean Regression, in which every instance is considered. In this way, the effect of mixed pixels cannot be eliminated. Though better than Prime-MIR, regressors cannot get the best prediction performance. When k is set to a proper value, such as 80 in 2020, the computed prototypes can preserve representative information and avoid the influence of mixed pixels. As a result, regressors can get the best prediction performance when k is set properly.

According to Table 1, the best k ranges from 60 to 80 in our experiments, and a more accurate k value may improve prediction performance. When k is too small, regressors cannot get enough information from prototypes. When k is too large, mixed pixels will affect the quality of prototypes, so it is essential to select a proper k value in VAEMIR. However, VAEMIR currently cannot automatically decide the best k value, so much work still needs to be done for VAEMIR. Though not perfect, VAEMIR is a promising method for large-scale corn yield prediction. Also, VAEMIR can be seen as a preprocessing method that can easily be combined with other ML algorithms for regression.

## V. Conclusions

We developed a VAE-based MIR method for large-scale corn yield prediction in this study. By adopting VAE for anomaly detection, we can preserve representative instances and avoid the influence of anomaly instances for multiple instance learning. Our experiments have proved that traditional MIR algorithms and mean regression cannot address the data variance caused by mixed pixels in a large-scale scene. Furthermore, VAEMIR can extract essential information from multiple instances and remove some noisy instances. As a result, VAEMIR outperforms all baseline methods in our corn yield prediction experiments. However, how to search best meta parameter k is still unaddressed in VAEMIR, which can significantly influence VAEMIR's performance. In the future, we will continue exploring how to automatically get the best k value in VAEMIR for better performance.